\documentclass[10pt,twocolumn,letterpaper]{article}

\usepackage{cvpr}              % To produce the CAMERA-READY version
\usepackage{multirow}
\usepackage{multicol}
\usepackage[accsupp]{axessibility} % Improves PDF readability for those with visual impairments.

\usepackage[dvipsnames]{xcolor}

\definecolor{cvprblue}{rgb}{0.21,0.49,0.74}
\usepackage[pagebackref,breaklinks,colorlinks,citecolor=cvprblue]{hyperref}

\def\paperID{15948} % *** Enter the Paper ID here
\def\confName{CVPR}
\def\confYear{2024}

\title{Bridging the Synthetic-to-Authentic Gap: Distortion-Guided Unsupervised Domain Adaptation for Blind Image Quality Assessment}

\author{Aobo Li \quad Jinjian Wu\thanks{Corresponding author.} \quad Yongxu Liu \quad Leida Li\\
Xidian University\\
{\tt\small abli@stu.xidian.edu.cn, jinjian.wu@mail.xidian.edu.cn, \{yongxu.liu, ldli\}@xidian.edu.cn}
}

\begin{document}
\maketitle
\begin{abstract}
The annotation of blind image quality assessment (BIQA) is labor-intensive and time-consuming, especially for authentic images. Training on synthetic data is expected to be beneficial, but synthetically trained models often suffer from poor generalization in real domains due to domain gaps. In this work, we make a key observation that introducing more distortion types in the synthetic dataset may not improve or even be harmful to generalizing authentic image quality assessment. To solve this challenge, we propose distortion-guided unsupervised domain adaptation for BIQA (DGQA), a novel framework that leverages adaptive multi-domain selection via prior knowledge from distortion to match the data distribution between the source domains and the target domain, thereby reducing negative transfer from the outlier source domains. Extensive experiments on two cross-domain settings (synthetic distortion to authentic distortion and synthetic distortion to algorithmic distortion) have demonstrated the effectiveness of our proposed DGQA. Besides, DGQA is orthogonal to existing model-based BIQA methods, and can be used in combination with such models to improve performance with less training data. 
\end{abstract}    
\section{Introduction}
\label{sec:intro}
Nowadays, countless images are being generated, transmitted, and processed by image enhancement, compression, and other algorithms all the time.
During these processes, distortions are inevitably introduced that degrade the image quality and may even affect the visual experience of receivers or the use of downstream algorithms. 
In most scenarios, the reference information required by full-reference (FR)  or reduced-reference (RR) image quality assessment (IQA) methods \cite{sun2018spsim, wu2013reduced} is not available. 
Therefore, it is crucial to have an accurate and efficient blind image quality assessment (BIQA) method \cite{ma2021blind} to monitor the image quality during these processes \cite{kim2017deep}.

\begin{figure}
    \includegraphics[width=\linewidth]{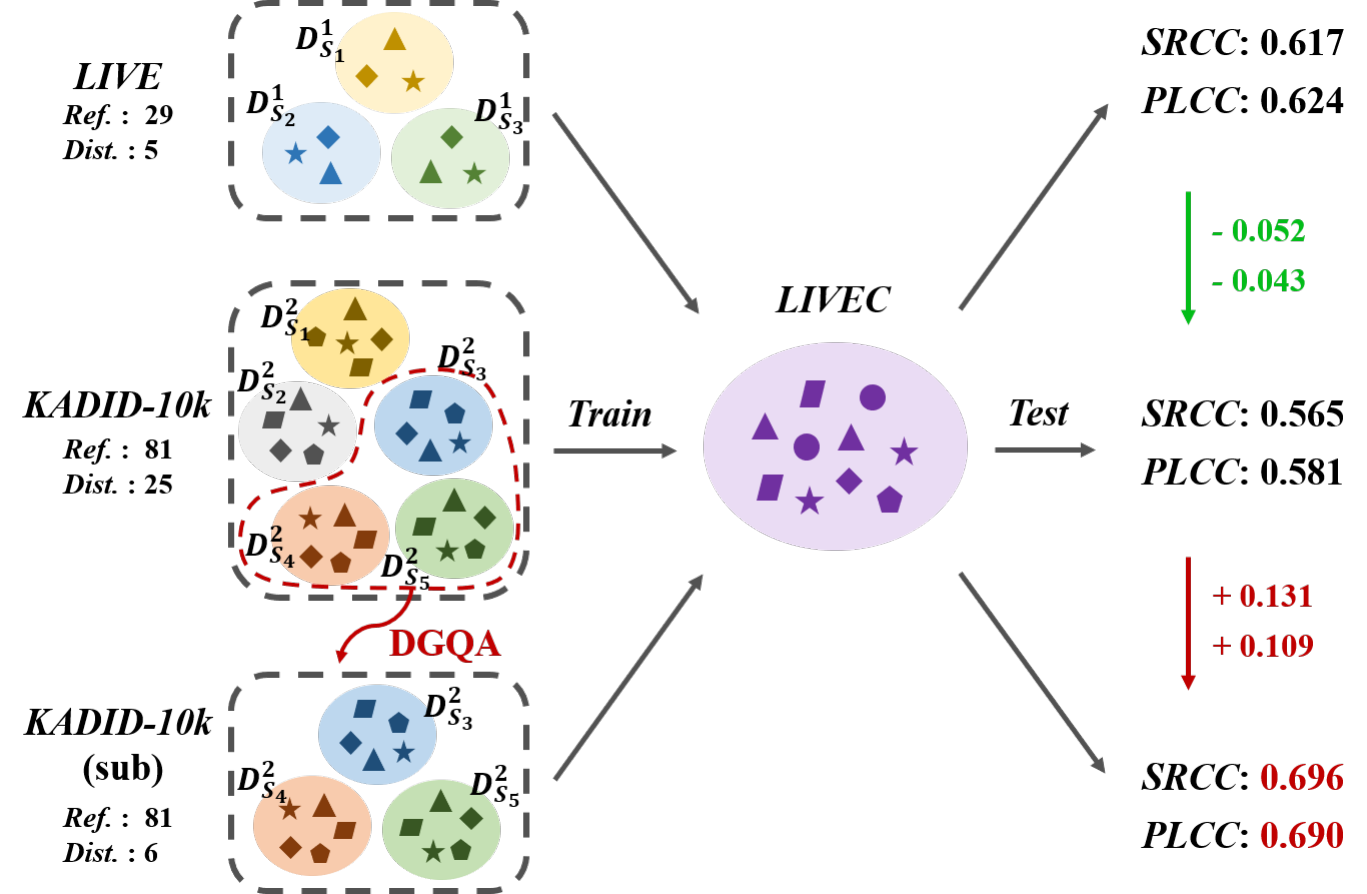}
    \caption{The performance on the authentically distorted dataset LIVEC \cite{ghadiyaram2015massive} when the baseline is trained on LIVE \cite{2006A}, KADID-10k \cite{kadid10k}, and the sub-set of KADID-10k selected by the proposed DGQA. Ref. and Dist. denote the number of the reference images and the number of the distortion in this dataset, respectively.} 
    \label{fig:performance_comparision}
\end{figure}

With the rapid development of deep learning, many learning-based BIQA methods \cite{kang2014convolutional, kim2016fully} have been proposed, which significantly improve the performance of traditional BIQA methods. 
However, since the annotation for the perceptual quality of images (especially authentic distortion images) is time-consuming and laborious, the amount of labeled data in existing BIQA datasets is relatively limited. It greatly limits the performance of deep learning-based BIQA methods that rely on a large number of labeled samples.

\begin{figure*}
    \includegraphics[width=\linewidth]{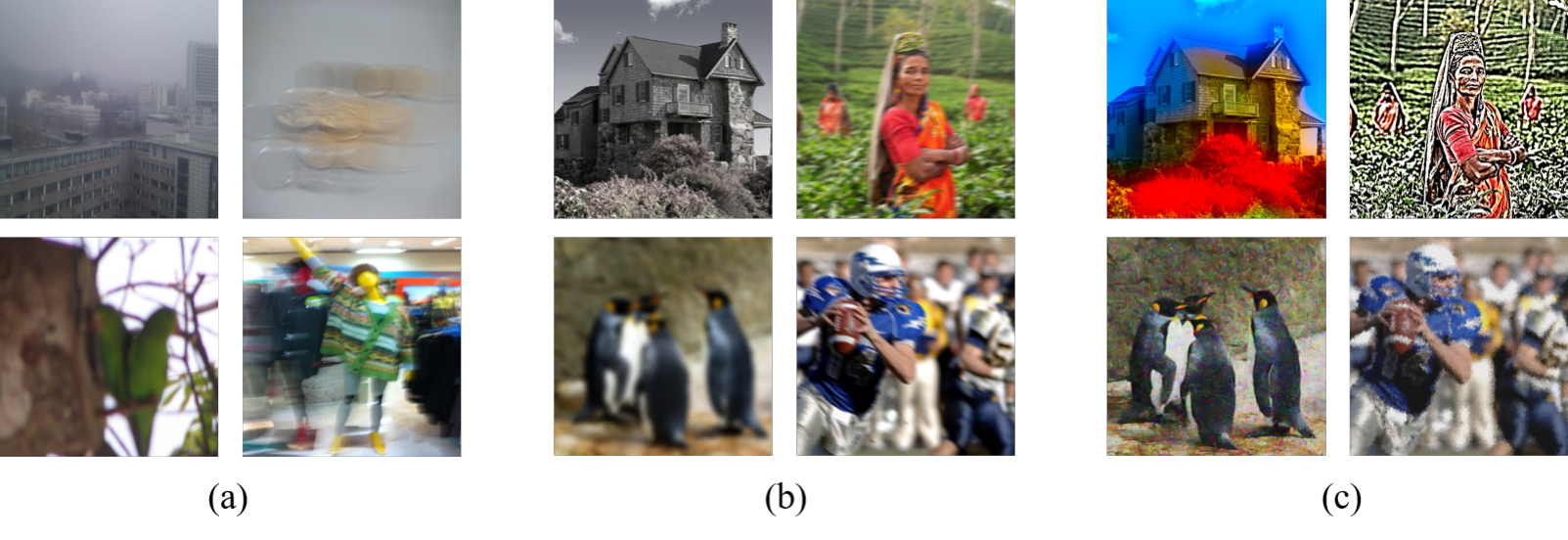}
    \caption{(a) are some representative images in LIVEC, (b) are some typical images in the part domains of KADID-10k whose styles are somewhat similar to that of LIVEC, and (c) are some typical images in the part domains of KADID-10k whose styles are quite different from LIVEC.} 
    \label{fig:example}
\end{figure*}

The acquisition of data and annotations can be much cheaper for synthetically distorted images generated by adding artificial distortions than for authentically distorted images, due to the availability of the corresponding reference images and the possibility of adjusting the parameters to generate different levels of distortion \cite{wang2023toward}.
If models trained on images with synthetic distortion could be effectively generalized to images with authentic distortion, it would greatly alleviate the pressure on annotation. 
However, since authentic distortions are complex and variable, 
it is difficult to simulate them directly by synthetic distortions.
As a result, there is a large gap between existing synthetic distortion datasets and authentic distortion datasets which makes existing methods perform poorly in syn-to-real generalization.

New synthetic distortion datasets introduce an increasing number of distortion types in an attempt to better cope with authentic distortion \cite{kadid10k}. 
However, we found that this did not actually lead to the expected performance improvement.
As shown in Fig. \ref{fig:performance_comparision}, compared to LIVE \cite{2006A}, the synthetic distortion dataset KADID-10k \cite{kadid10k} with more distortion types and more reference images surprisingly exhibits significantly worse generalization performance on the authentic distortion dataset LIVEC \cite{ghadiyaram2015massive}.

Generalizing a model from a synthetically distorted dataset with multiple artificial distortions to an unlabeled authentically distorted dataset can be modeled as an unsupervised multi-source domain adaptation (UMDA) problem.
As shown in Fig. \ref{fig:example}, we can observe that the styles of some domains in the synthetically distorted dataset are often somewhat similar to that of the synthetically distorted dataset, while some are quite different.
Some studies on domain adaptation (DA) \cite{zhao2018adversarial, cao2018partial} have shown that outlier source domains that differ too much from the target domain will may not result in performance gains or even negative transfers.
Therefore, a natural idea is to try to sieve out the outlier domains from the multi-source domains, while retaining the source domains that are similar to the target domain. 

In this work, we analyze how to measure the distance of source and target domains for BIQA under unsupervised scenarios and propose a distortion-guided unsupervised domain adaptation framework for BIQA (DGQA). 
By performing similar domain selection according to the target domain, source domains whose distributions differ significantly from the target domain distribution are filtered out. The filtered similar source domains are used for training the model, which reduces the negative transfer from outlier source domains to the target domain and improves the generalization ability of the model.
As shown in Fig. \ref{fig:performance_comparision}, our framework DGQA improves by $13.1\%$ in SRCC and $10.1\%$ in PLCC when tested on LIVEC using only $24\%$ of the training data. Our contribution is summarized as follows:

\begin{itemize}
    \item[$\bullet$] We make a key observation that the use of a larger synthetic distortion dataset may not improve or even be harmful to the performance of model generalization to authentic distortions, and analyze the reasons for this in the context of some research about domain adaptation.
    \item[$\bullet$] We analyze how to measure the distance of source and target domains for BIQA under unsupervised scenarios and accordingly propose a distortion-guided unsupervised domain adaptation framework that reduces the negative transfer of the outlier source domains to the target domain and improves the generalization capacity of the model.
    \item[$\bullet$] Our approach significantly improves the model performance with a much smaller amount of training data in the settings of synthetic distortion to authentic distortion and synthetic distortion to algorithmic distortion. Besides, since our method is a sample-based approach, it is fully compatible with existing model-based approaches.
\end{itemize}

\section{Related Work}
\label{sec:related_work}
In this section, we give a brief overview of recent progress in BIQA. We then review unsupervised domain adaptation (UDA) and its applications for BIQA.

\subsection{Blind Image Quality Assessment}
Early traditional methods typically use handcrafted features based on natural scene statistics \cite{mittal2012no, mittal2012making, venkatanath2015blind}. They extract these features for images and map them to quality scores to achieve BIQA. Due to the complexity of the BIQA task and the limited ability to hand-crafted feature representations, these methods generally fail to perform satisfactorily. 

In recent times, many BIQA approaches \cite{kang2014convolutional, kim2016fully, liu2017rankiqa} based on deep learning have been proposed. These approaches have made remarkable progress beyond traditional methods, as the deep neural network possesses exceptional ability for representation learning. 
Zhang et al. \cite{zhang2018blind} design DB-CNN consisting of two pre-trained convolutional neural networks which extract features about synthetic and authentic distortions, respectively.
MetaIQA \cite{zhu2020metaiqa} utilizes meta-learning to learn prior knowledge that is shared amongst diverse types of distortions.
In \cite{su2020blindly}, Su et al. propose a self-adaptive hypernetwork for BIQA. The model first extracts semantic features from images, subsequently employs a hypernetwork to dynamically learn perception rules, and finally predicts the quality of the images.
Zhao et al. \cite{zhao2023quality} improve the existing degradation process and propose a self-supervised learning framework for BIQA. 
Qin et al. \cite{qin2023data} present a new transformer-based BIQA method that efficiently learns quality-aware features with minimal data and introduces a human-inspired attention panel mechanism.

However, these methods are only applicable in scenarios where the training and testing data are from similar distributions and exhibit poor performance when trained on the synthetic distortion dataset and tested on the authentic distortion dataset since there is a large gap between synthetic distortions and authentic distortions.

\subsection{Unsupervised Domain Adaptation}
Unsupervised domain adaptation (UDA) aims to align the distribution of the labeled source domain and the unlabeled target domain, thereby improving the generalization performance from the source domain to the target domain.

The majority of UDA approaches are feature-based, which can be classified into two categories: discrepancy-based methods and adversary-based methods. 
The former primarily employ distance measures such as Wasserstein metric, Kullback-Leibler (KL) divergence \cite{kullback1951information}, maximum mean discrepancy (MMD) \cite{gretton2006kernel}, contrastive domain discrepancy (CDD) \cite{kang2019contrastive}, and correlation alignment (CORAL) \cite{sun2017correlation} to assess the similarity between domains.
The latter aims to learn transferable and domain-invariant features using the domain discriminator to encourage domain confusion through an adversarial objective, like DANN \cite{ajakan2014domain}, and CDAN \cite{long2018conditional}.

Currently, there are a few works that introduce UDA into BIQA. UCDA \cite{chen2021unsupervised} divides the target domain into confident and non-confident subdomains and subsequently aligns the source domain and target subdomains from easier to more challenging. Chen et al. \cite{chen2021no} apply the center loss to acquire domain discriminative features for different domains. They subsequently aligned two domains using ranked paired features to facilitate knowledge transfer from natural IQA to screen content IQA. Lu et al. \cite{lu2022styleam} utilize the feature style space for BIQA as the alignment space and propose Style Mixup to improve the consistency between the feature styles and their quality scores.

However, BIQA is oriented to images with very large differences in style, and imposing conventional feature-based UDA methods can lead to severe negative transfer due to the huge gaps between domains. For this reason, we start from the data level and filter the source domains that are more similar to the target domain among the set of source domains to align the source and target domains.

\section{Proposed Method}
\label{sec:method}
\begin{figure*}
    \centering
    \includegraphics[width=\linewidth]{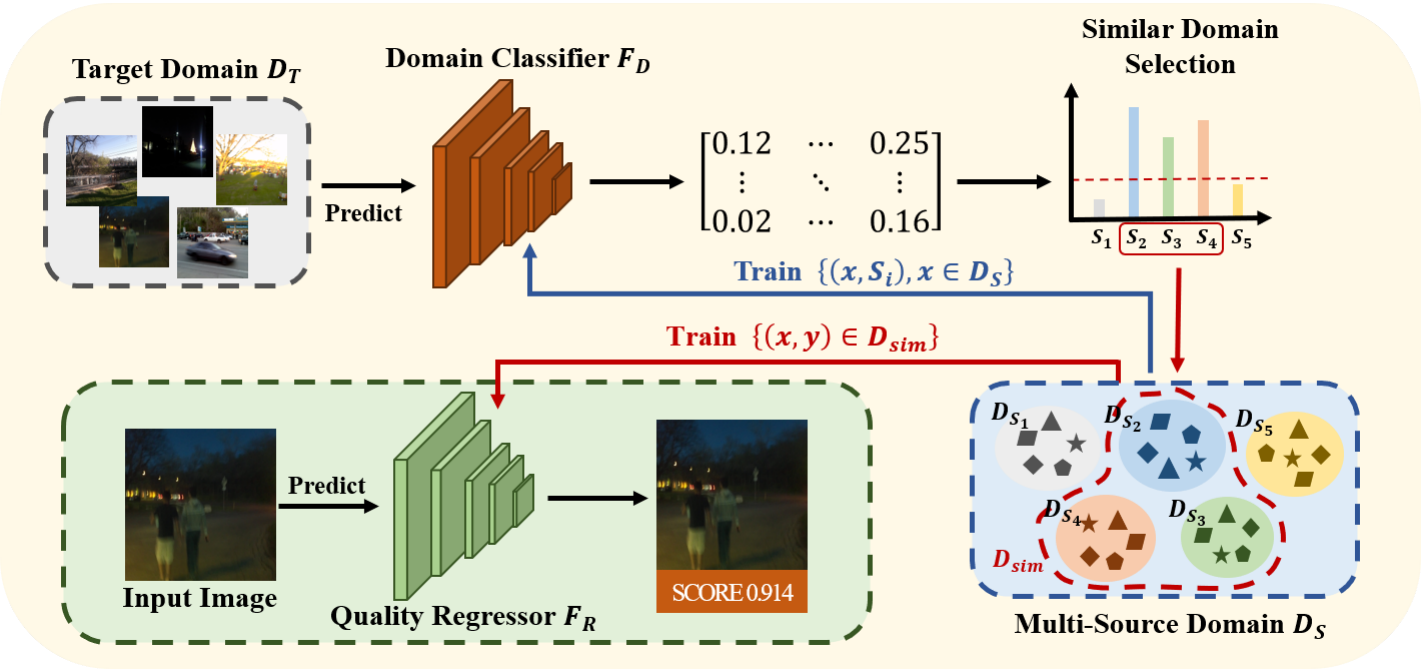}  
    \caption{The framework of the proposed DGQA. By performing similar domain selection, the source domains with a large gap to the target domain are sieved out. The sieved similar domains $\mathcal{D}_{sim}$ are used to train the model, which reduces the negative transfer from outlier source domains to the target domain and improves the model's performance on the target domain.} 
    \label{fig:framework}
\end{figure*}

In this section, we first introduce the preliminaries of the problem formulation and the factors affecting the generalization error bound of the target domain. Then we describe how to implement distortion-guided source domain selection. Finally, we present the training strategy of our proposed framework. The framework of the proposed DGQA is shown in Fig. \ref{fig:framework}.

% 1. （直接先将问题形式化，先说合成数据集泛化问题可以看做是一个多源域自适应的问题，然后用公式描述问题，随后再导出泛化误差上界）

\subsection{Preliminary}
The problem of generalizing a synthetically distorted IQA dataset containing $k$ types of artificial distortions to a specific real scenario can be viewed as an unsupervised multi-source domain adaptation (UMDA) problem. For this problem, we have $k$ labeled source domains $\{\mathcal{D}_{S_i}=\{x_{S_i}^j, y_{S_i}^j\}^n_{j=1}\}^k_{i=1}$ generated based on different distortion types and an unlabeled target domain $\mathcal{D}_T=\{x_T^j\}^n_{j=1}$, where $x_{S_i}^j$ and $y_{S_i}^j$ denote the images and the corresponding quality scores of the source domains, and $x_T^j$ denotes the images of the target domain, respectively. We expect to achieve good performance on the target domain by learning on these source domains that have different degrees of gap with the target domain.

The generalization error bound of the model over the target domain can be used as a measure of the performance of the task in the target domain, and \cite{zhao2018adversarial} gives the bound for multi-source domain adaptation:

\textbf{Theorem 1.} Let $\mathcal{H}$ be a set of real-valued functions from $X$ to $[0, 1]$ with $P\textit{dim}(\mathcal{H}) = d$. If $\{\hat{\mathcal{D}}_{S_i}\}^k_{i=1}$ are the empirical distributions generated with $\textit{m}$ \textit{i.i.d.} samples from each domain, and $\hat{\mathcal{D}}_T$ is the empirical distribution on the target domain generated from $mk$ samples without labels, then,  ${\forall}\alpha \in R^k_+, \sum_{i \in [k]}{\alpha_i} = 1$, and for $0 < \delta < 1$, with probability at least $1-\delta$, for all $h \in H$, we have:
\begin{align}
    \epsilon_T(h) & \leq 
                    \sum_{i \in [k]}{\alpha_i}
                    \left (\hat{\epsilon}_{S_i}(h)+\frac{1}{2} d_{\overline{\mathcal{H}}}(\hat{\mathcal{D}}_T;\hat{\mathcal{D}}_{S_i})\right )  
                    + \lambda_\alpha \nonumber \\
                  & + \mathcal{O} \left (\sqrt{\frac{1}{km}(log\frac{1}{\delta}+dlog\frac{km}{d})} \right )
    \label{eq:important}
\end{align}

\begin{equation}
    d_{\overline{\mathcal{H}}}(\hat{\mathcal{D}}_T;\hat{\mathcal{D}}_{S_i})
    =2 \sup_{\overline{h} \in \overline{\mathcal{H}}}{|\Pr_{\hat{\mathcal{D}}_{S_i}}(h)-\Pr_{\hat{\mathcal{D}}_T}}(h)|
    \label{eq.2}
\end{equation}
where $\epsilon_T(h)$ is the generalization error of the target domain, $\hat{\epsilon}_{S_i}(h)$ is the empirical error in the source domain $S_i$, $d_{\overline{\mathcal{H}}}(\hat{\mathcal{D}}_T;\hat{\mathcal{D}}_{S_i})$ is the difference in the empirical distributions of the source domain $S_i$ and the target domain $T$, $\lambda_\alpha$ is the risk of the optimal hypothesis on the mixture source domain $\sum_{i \in [k]}{\alpha_{i}S_{i}}$ and $T$, and $\overline{\mathcal{H}}:=\{\mathbb{I}_{|h(x)-h'(x)|>t}:h,h'\in \mathcal{H},0\leq t\leq 1\}$ is the set of threshold functions induced from $\mathcal{H}$.

According to Thm. 1, we can know that the factors affecting the generalization error bound of the target domain are source domain empirical error, distance between the distributions of the source domains and the target domain, optimal joint generalization error, and sample size, respectively. Since the empirical error can be minimized to an arbitrarily small value by training, the optimal joint generalization error does not depend on a specific hypothesis $h$ \cite{ajakan2014domain}. Therefore, besides adding training samples, we can reduce the generalization error of the model on the target domain by narrowing the distance between the distributions of the source domains and the target domain, thus improving the performance of the target domain \cite{dou2019domain}.

\subsection{Distortion-Guided Source Domain Selection}
BIQA is oriented to images with very large differences in style, and imposing conventional feature-based domain adaptive methods can lead to severe negative transfer due to the huge gaps between domains. 
For this reason, we start from the data level and filter the source domains that are more similar to the target domain among the set of source domains to reduce the negative transfer caused by outlier source domains.

For BIQA, the data distribution is determined by the content distribution $\mathcal{D}^c$  and distortion distribution $\mathcal{D}^d$ \cite{li2022blind}. 
Since images with any content can be selected as the reference images to generate synthetic distortion images, it is theoretically possible to achieve a content distribution that is very similar to an authentic distortion image dataset. Hence, we assume that the source domain has the same image content as the target domain, i.e. $\mathcal{D}^c_{S_i} = \mathcal{D}^c_T$. At this point, the distance between the distributions $\mathcal{D}$ of the source domains and the target domain can be translated into the distance between the distortion distribution $\mathcal{D}^d$ of the source domains and the target domain.
In addition, all source domains are sampled uniformly, i.e. $\alpha_i=\frac{1}{k}$. Then the distance between the distributions of the source domains and the target domain can be expressed as
\begin{equation}
    d_{\overline{\mathcal{H}}}(\hat{\mathcal{D}}_T;\hat{\mathcal{D}}_{S_i})
    =\frac{2}{k} \sum_{i \in [k]} {\sup_{\overline{h} \in \overline{\mathcal{H}}}{|\Pr_{\hat{\mathcal{D}}^d_{S_i}}(h)-\Pr_{\hat{\mathcal{D}}^d_T}}(h)|}
\end{equation}
This distance can be interpreted as the discriminator's discriminability of source and target domain distortions.

Since our goal is only to filter the source domains that are relatively more similar to the target domain, for this reason, we only need to compare the relative discriminability of each source domain to the target domain.
To this end, we integrate the domain discriminators for multiple source domains into a single multi-source domain classifier $F_D$.
For each sample $x^T_j$ in the target domain, input the multi-source domain classifier $F_D$ to predict the probability $p$ that the sample belongs to each source domain, 
\begin{equation}
    [p^1_j, p^2_j, ..., p^k_j] = F_D(x^T_j)
\end{equation}
The relative similarity between the target domain $\mathcal{D}_T$ and the source domain $\mathcal{D}_{S_i}$ is defined as the average probability that all the images in the target domain belong to this source domain, 
\begin{equation}
    sim(\mathcal{D}_T, \mathcal{D}_{S_i}) = \frac{1}{N}\sum_{j=1}^N p^i_j
\end{equation}
where $N$ is the simple size in the target domain.
Finally, the source domains with relative similarity above the threshold $\tau$ ($\tau$ is set as $\frac{1}{k}$ in this work) are taken as the similar domains of the target domain, 
\begin{equation}
    \mathcal{D}_{sim} = \{\mathcal{D}_{S_j}| sim(\mathcal{D}_T, \mathcal{D}_{S_j}) > \tau, j \in [k]\}
\end{equation}
Through the distortion-guided source domain selection, source domains with significantly large differences in distribution from the target domain are filtered out, thus reducing the negative transfer of outlier source domains to the target domain and improving the generalization ability of the model.

\subsection{Training}
\textbf{Domain Classify.} 
We choose the 25 classes of distortion types in KADID-10k as the domain codes. 
The tailored ResNet-50 whose fully connected (FC) layer is replaced with an FC layer with 25 nodes and a softmax layer is the multi-source domain classifier $F_D$. 
To get better classification ability, KADIS-700k, which has a larger data size and the same distortion types as KADID-10k, is used to train $F_D$. Cross entropy is the loss function used for domain classification, 
\begin{equation}
    L_D = \sum^M_{j=1} (-y^D_j \log(F_D(x_j)) - (1-y^D_j) \log(1-F_D(x_j)))
\end{equation}
where $y^D_j$ is the domain encoding of $x_j$, and $M$ is the sample size of train data.  

\noindent\textbf{IQA Regression.}
Similarly, the tailored ResNet-50 whose FC layer is replaced with an FC layer with one node is the IQA regressor $F_R$. As shown in Fig. \ref{fig:framework}, the selected similar source domains are used as training data for training the model. We use the L1 norm as the loss function for image quality score prediction, 
\begin{equation}
    L_R = \sum_{(x_j, y_j) \in \mathcal{D}_{sim}} ||F_R(x_j) - y_j||_1
\end{equation}

\section{Experiments}
\begin{table*}[]
\caption{Summary of the datasets for training or evaluation.}\label{tab:db_summary}
\centering
\begin{tabular}{l|cc|ccc|c}
\hline
        Dataset           & KADID-10k  & KADIS-700k   & LIVEC        & BID        & KonIQ-10k  & PIPAL     \\ 
        \hline
        Reference images  & 81         & 140000       & N/A          & N/A        & N/A        & 250       \\ 
        Distorted images  & 10125      & 700000       & 1162         & 586        & 10073      & 29000\\ 
        Distortion types  & 25         & 25           & N/A          & N/A        & N/A        & 40        \\ 
        Annotation        & DMOS       & N/A          & MOS          & MOS        & MOS        & MOS       \\ 
        Scenario          & Synthetic  & Synthetic    & Authentic    & Authentic  & Authentic  & Synthetic+Algorithmic \\ 
\hline
\end{tabular}
\label{sec:experiments}
\end{table*}

In this section, we first describe the experimental setup in detail. Then the performance comparison is made with the several state-of-the-art methods and UDA methods on the synthetic-to-authentic and synthetic-to-algorithmic scenario. A series of subsequent experiments validate the generalisability of our method, the effectiveness of the proposed core components, and the compatibility with model-based methods. Finally, we explore the upper bound of the performance of the method based on source domain selection on KADID-10k \cite{kadid10k} for generalizing to authentic distortion datasets and experimentally demonstrate the feasibility of generalizing to authentic distortions by generalizing over synthetic distortions.

\begin{table*}
    \centering
    \caption{Similar source domain statistics for all target domains, where N.o.S. denotes the number of similar source domains for this target domain, and N.o.T. denotes the number of occurrences of this source domain.} 
    \begin{tabular}{l|ccccccccccccccccc|c}
    \hline
        ~ & \#1 & \#2 & \#3 & \#4 & \#7 & \#9 & \#10 & \#11 & \#12 & \#16 & \#17 & \#18 & \#19 & \#21 & \#22 & \#23 & \#25 & N.o.S  \\ \hline
        LIVEC & \checkmark & ~ & \checkmark & ~ & ~ & ~ & \checkmark & ~ & ~ & ~ & \checkmark & \checkmark & ~ & ~ & ~ & ~ & \checkmark & 6  \\ 
        Koniq-10k & \checkmark & ~ & \checkmark & ~ & ~ & ~ & ~ & ~ & ~ & \checkmark & \checkmark & \checkmark & ~ & ~ & ~ & ~ & \checkmark & 6  \\ 
        BID & \checkmark & ~ & \checkmark & ~ & ~ & \checkmark & ~ & ~ & ~ & ~ & ~ & ~ & ~ & \checkmark & ~ & \checkmark & \checkmark & 6  \\ \hline
        PIPAL-1 & \checkmark & \checkmark & \checkmark & ~ & ~ & \checkmark & ~ & ~ & ~ & ~ & \checkmark & \checkmark & ~ & ~ & ~ & ~ & \checkmark & 7  \\
        PIPAL-2 & \checkmark & \checkmark & \checkmark & ~ & ~ & \checkmark & ~ & ~ & ~ & ~ & \checkmark & ~ & ~ & ~ & ~ & \checkmark & ~ & 6  \\ 
        PIPAL-3 & \checkmark & ~ & \checkmark & ~ & ~ & ~ & ~ & ~ & ~ & \checkmark & \checkmark & \checkmark & \checkmark & ~ & ~ & ~ & \checkmark & 7  \\ 
        PIPAL-4 & \checkmark & \checkmark & \checkmark & ~ & \checkmark & \checkmark & \checkmark & ~ & ~ & ~ & \checkmark & \checkmark & ~ & ~ & ~ & ~ & \checkmark & 9  \\ 
        PIPAL-5 & \checkmark & \checkmark & \checkmark & \checkmark & ~ & \checkmark & ~ & ~ & ~ & ~ & ~ & ~ & ~ & ~ & ~ & ~ & ~ & 5  \\ 
        PIPAL-6  & \checkmark & \checkmark & \checkmark & ~ & ~ & \checkmark & \checkmark & \checkmark & \checkmark & \checkmark & \checkmark & \checkmark & \checkmark & \checkmark & \checkmark & ~ & \checkmark & 14  \\ \hline
        N.o.T & 9 & 5 & 9 & 1 & 1 & 6 & 3 & 1 & 1 & 3 & 7 & 6 & 2 & 2 & 1 & 2 & 7 &   \\ 
    \hline
    \end{tabular}
    \label{tab:DGDS_ana}
\end{table*}

\subsection{Experimental Setup}
We conduct experiments on six IQA datasets, containing the synthetic distortion datasets KADID-10k \cite{kadid10k}, KADIS-700k \cite{kadid10k}, the authentic distortion datasets LIVEC \cite{ghadiyaram2015massive}, BID \cite{ciancio2010no}, KonIQ-10k \cite{hosu2020koniq}, and the dataset PIPAL \cite{pipal} with both synthetic and algorithmic distortions. 
The specific information of all datasets is summarised in Table \ref{tab:db_summary}. 
Two typical criteria, SRCC and PLCC are adopted for measuring prediction monotonicity and prediction accuracy. 

Specifically, KADIS-700k is randomly divided into 80\% training data and 20\% testing data and used for the pre-training of the domain classifier. 
KADID-10k is also partitioned into training and validation sets in the same proportions for the training of the IQA regressor, and the remaining four datasets are used as test sets respectively. 
All splitting is performed by reference images to ensure no overlapping content. 
This random train-val splitting on KADID-10k for all experiments is repeated five times and the median SRCC and PLCC are reported. 

During the training process, we randomly sample a patch of size $224\times224\times3$ from each training image, and random horizontal flipping is used for data augmentation. Our mini-batch size is set at 32, while the learning rate is at $2\times10^{-5}$. We use the Adam \cite{kingma2014adam} optimizer coupled with a weight decay of $5\times10^{-4}$ to optimize the model for 15 epochs. In the testing phase, five patches from each image are randomly sampled and the mean of their prediction results is the final output. This method is implemented by Pytorch, and all experiments are carried out on NVIDIA TITAN Xp GPUs.

\subsection{Analysis of Similarity Source Domain}
For the target domains (LIVEC, KonIQ-10k, BID, and the algorithmically distorted part of PIPAL) that are used as test sets in the experiments, we use distortion-guided domain selection (DGDS) to filter their similar source domains in the training set KADID-10k. The results are shown in Table \ref{tab:DGDS_ana}, where only source domains with at least 1 occurrence are listed. PIPAL-1, 2, 3, 4, 5, 6 denote traditional SR, PSNR-originated SR, SR with kernel mismatch, GAN-based SR, denoising, SR and denoising joint in PIPAL, respectively. 

From Table \ref{tab:DGDS_ana}, we can see that for most target domains, similar source domains account for less than 1/3 of the total source domains. And based on subsequent experimental results, we will find that the remaining source domains do not lead to performance improvement, and may even seriously affect the performance of the model in most cases.

In addition, \#1 (Gaussian blur), \#3 (Motion blur), \#9 (JPEG2000), \#17 (Darken), \#18 (Mean shift), and \#25 (Contrast change) can be found in almost all the similar source domains of the target domains. It might provide some reference for the selection of source domains for unseen target domains.

These filtered similar source domains will be used in subsequent experiments for the training of the IQA regressor (denoted DGQA). The IQA regressor trained on all source domains (i.e., the complete KADID-10k) serves as the baseline. Besides, the compared methods are trained on all source domains by default, when not specifically noted.

\subsection{Performance Evaluation}
\textbf{1) Performance on the synthetic-to-authentic setting.}
On this setting, the proposed DGQA is trained on KADID-10k and tested on LIVEC, KonIQ-10k, and BID. The performance of DGQA is compared against ten BIQA approaches, including three traditional methods  (BRISQUE \cite{mittal2012no}, NIQE \cite{mittal2012making} and PIQE \cite{venkatanath2015blind}), three deep learning-based approaches (RankIQA \cite{liu2017rankiqa}, MetaIQA \cite{zhu2020metaiqa}, DB-CNN \cite{zhang2018blind} and  HyperIQA \cite{su2020blindly}) and four UDA-based methods (DANN \cite{ajakan2014domain}, UCDA \cite{chen2021unsupervised}, RankDA \cite{chen2021no} and StyleAM \cite{lu2022styleam}). The comparisons are summarized in Table \ref{tab:cross_authentic} and the bolded results imply top performance. All comparison results are from the publicly available paper \cite{lu2022styleam}.

\begin{table*}[]
\centering
\caption{Performance comparison on the synthetic-to-authentic setting (KADID-10k$\rightarrow$LIVEC, KonIQ-10k, and BID). The average results are in the last column.
}\label{tab:cross_authentic}
\begin{tabular}{l|cc|cc|cc|cc}
\hline
Methods         & \multicolumn{2}{c|}{LIVEC}     & \multicolumn{2}{c|}{KonIQ-10k}       & \multicolumn{2}{c|}{BID}            & \multicolumn{2}{c}{Average} \\
                & SRCC&PLCC                       & SRCC&PLCC                            & SRCC&PLCC                            & SRCC&PLCC  \\
\hline
BRISQUE         &0.2433&0.2512                    &0.1077&0.0991                         & 0.1745&0.1750                        & 0.1752&0.1751\\
NIQE            &0.3044&0.3619                    &0.4469&0.4600                         & 0.3553&0.3812                        & 0.3689&0.4010 \\
PIQE            &0.2622&0.3617                    &0.0843&0.1995                         & 0.2693&0.3506                        & 0.2053&0.3039 \\
\hline
RankIQA         &0.4906&0.4950                    &0.6030&0.5511                         & 0.5101&0.3671                        & 0.5346&0.4711 \\
MetaIQA         &0.4639&0.4638                    &0.5006&0.5035                         & 0.3005&0.4280                        & 0.4217&0.4651 \\
DBCNN           &0.2663&0.2897                    &0.4126&0.4209                         & 0.3179&0.2115                        & 0.3323&0.3074 \\
HyperIQA        &0.4903&0.4872                    &0.5447&0.5562                         & 0.3794&0.2820                        & 0.4715&0.4418 \\
\hline
DANN            &0.4990&0.4835                    &0.6382&0.6360                         & 0.5861&0.5102                        & 0.5744&0.5432 \\
UCDA            &0.3815&0.3584                    &0.4958&0.5010                         & 0.3480&0.3907                        & 0.4084&0.4167 \\
RankDA          &0.4512&0.4548                    &0.6383&0.6227                         & 0.5350&0.5820                        & 0.5415&0.5532 \\
StyleAM         &0.5844&0.5606                    &\textbf{0.7002}&0.6733                & 0.6365&0.5669                        & 0.6404&0.6003 \\
\hline
DGQA        & \textbf{0.6958}&\textbf{0.6902}     & 0.6810&\textbf{0.6866}      & \textbf{0.7696}&\textbf{0.7526}    & \textbf{0.7155}&\textbf{0.7098} \\

\hline
\end{tabular}
\end{table*}

As can be seen, the proposed DGQA uses less than 1/3 of the training data and achieves SOTA performance when generalized to all authentic distortion datasets. Only StyleAM is competitive on the "KADID-10k$\rightarrow$KonIQ-10k" scenario. In all other scenarios, our method achieves significant improvements (SRCC increases more than $11.14\%$ when tested on LIVEC, $13.31\%$ when tested on BID) and the average SRCC and PLCC increase more than $7.51\%$ and $10.95\%$, respectively.  
This demonstrates that our DGQA framework, which significantly reduces the negative transfer from the outlier domains by reduces the distance between the target domain and the source domains, and thus effectively improves the model's performance in generalizing from synthetic distortion to authentic distortion.

\noindent\textbf{2) Performance on the synthetic-to-algorithmic setting}. To further explore the universality of our proposed framework, we conduct experiments on the "KADID-10k$\rightarrow$algorithmic distortions on PIPAL" scenario. The experimental results are shown in Table \ref{tab:cross_algorithmic}. 
Our proposed framework improves the performance of baseline on all algorithmic distortion types in PIPAL except denoising, and SRCC improves more than $5\%$ on four algorithmic distortion types including traditional SR, PSNR-originated SR, SR, and denoising joint and even on the very challenging GAN-based SR. Although our performance on denoising is slightly lower than the baseline, it should be noted that it only has used 1/5 labeled data compared to the baseline.
The above experimental results show that our method is not only effective in generalizing synthetic distortions to authentic distortions but also helps to improve the performance of generalization to other distortions.

\begin{table*}[]
    \centering
    \caption{Performance comparison on the synthetic-to-algorithmic setting (KADID-10k$\rightarrow$algorithmic distortions on PIPAL). The average results are in the last row.} \label{tab:cross_algorithmic}
    \begin{tabular}{l|cc|cc}
    \hline
        \multirow{2}{*}{Sub-type}& \multicolumn{2}{c|}{Baseline} & \multicolumn{2}{c}{DGQA}  \\
        ~ & SRCC & PLCC & SRCC & PLCC  \\ \hline
        Traditional SR          & 0.4756  & 0.4536  & $\textbf{0.5419}_{+6.63\%}$  & $\textbf{0.5404}_{+8.68\%}$   \\
        PSNR-originated SR      & 0.4758  & 0.4697  & $\textbf{0.5810}_{+10.52\%}$ & $\textbf{0.5956}_{+12.59\%}$   \\
        SR with kernel mismatch & 0.1609  & 0.1121  & $\textbf{0.1629}_{+0.20\%}$  & $\textbf{0.1353}_{+2.32\%}$   \\ 
        GAN-based SR            & 0.4898  & 0.4840  & $\textbf{0.5393}_{+4.95\%}$  & $\textbf{0.5279}_{+4.39\%}$   \\
        Denoising               & \textbf{0.5705}   & \textbf{0.5285}  & $0.5588_{-1.17\%}$   & $0.5193_{-0.92\%}$   \\
        SR and Denoising Joint  & 0.3941  & 0.3836  & $\textbf{0.4470}_{+5.29\%}$  & $\textbf{0.4390}_{+5.54\%}$   \\ \hline
        Average	                & 0.4278   & 0.4053 & $\textbf{0.4718}_{+4.40\%}$  & $\textbf{0.4596}_{+5.43\%}$   \\ \hline
    \end{tabular}
\end{table*}

\subsection{Ablation Study}
To explore the contribution of the key component and the generalisability to different models of our DGQA, a series of ablation experiments are conducted in this subsection.
The baseline (ResNet-50), HyperIQA and StyleAM trained on complete KADID-10k are each used as a benchmark to test the performance gain of our approach. 
The experimental results are shown in Table \ref{tab:ablation}, where the results of HyperIQA are obtained by the source code released by the original authors, and those of StyleAM are obtained by our reproduced code. 

\begin{table*}[]
\centering
\caption{Ablation studies on the key component and different models.}\label{tab:ablation}
\begin{tabular}{l|c|ll|ll}
\hline
\multirow{2}{*}{Methods}         & \multirow{2}{*}{Domain}     & \multicolumn{2}{c|}{LIVEC}                             & \multicolumn{2}{c}{KonIQ-10k} \\
                                 &                           & SRCC&PLCC                                                & SRCC  &PLCC                       \\
\hline
\multirow{2}{*}{Baseline}        & All                      &0.5646&0.5810                                              &0.6250 &0.6127\\
                                 & DGDS                     &$\textbf{0.6958}_{+13.12\%}$ & $\textbf{0.6902}_{+10.92\%}$           &$\textbf{0.6810}_{+5.60\%}$&$\textbf{0.6866}_{+7.39\%}$\\
\hline
\multirow{2}{*}{HyperIQA}        & All                      &0.5633&0.5890                                                          &0.6236 &0.5869      \\
                                 & DGDS                     &$\textbf{0.6944}_{+13.11\%}$ & $\textbf{0.6974}_{+10.84\%}$            &$\textbf{0.6853}_{+6.17\%}$ &$\textbf{0.6628}_{+7.59\%}$      \\
\hline
\multirow{2}{*}{StyleAM}         & All                      &0.5174&0.5219                                                          &0.7046 &0.7060      \\
                                 & DGDS                     &$\textbf{0.6687}_{+15.13\%}$ & $\textbf{0.6671}_{+14.52\%}$            &$\textbf{0.7088}_{+0.43\%}$ &$\textbf{0.7263}_{+2.03\%}$      \\
\hline
\end{tabular}
\end{table*}

It can be observed that our method has significantly improved the performance of the baseline by DGDS. The SRCC and PLCC improve $13.12\%$, $10.92\%$ when tested on LIVEC, and $5.60\%$, $7.39\%$ when tested on KonIQ-10k. It shows the effectiveness of the distortion-guided domain selection.
Besides, similar improvements in performance can be seen on HyperIQA and StyleAM. This verifies the generalisability to different models and the compatibility with model-based approaches of our DGQA. Besides, the performance of the combination of StyleAM and DGDS on Koniq-10k shows the potential of combining our method with feature-based UDA methods to further improve performance.

\subsection{Further Study}
In this section, we further explore the upper bound of the performance on generalization to authentic datasets based on source domain selection on KADID-10k and compare it with the performance on generalization from an authentic dataset to other authentic datasets. The performance comparisons are shown in Table \ref{tab:further}.

Specifically, we adopt a greedy source domain selection (GDS) strategy to incrementally introduce the source domains in KADID-10k. When tested on LIVEC, the set of source domains consisting of \#1 (Gaussian blur), \#3 (Motion blur), \#9 (JPEG2000), \#18 (Mean shift), \#20 (Non-eccentricity patch), and \#25 (Contrast change) performs best. And when tested on KonIQ-10k, the source domains consisting of \#1 (Gaussian blur), \#2 (Lens blur), \#9 (JPEG2000), \#17 (Darken), \#20 (Non-eccentricity patch), and \#25 (Contrast change) perform best.
We can observe that the set of source domains selected by GDS is similar to ours, which validates the effectiveness of our distortion-directed source domain selection.

Furthermore, while there is still a large gap compared to the performance of KonIQ-10k generalized to LIVEC, it is worth noting that the amount of data for the set of source domains selected via the greedy selection strategy is just less than 1/6 of that of KonIQ-10k.
Compared to LIVEC, which is of the same magnitude, an even higher SRCC is achieved for the set of source domains selected via GDS when generalized to KonIQ-10k.
This demonstrates the feasibility of generalizing to authentic distortions by generalizing over synthetic distortions.

\begin{table}[]
\centering
\caption{Performance comparison on generalization to authentic datasets from the source domain set selected on KADID-10k and other authentic datasets.}\label{tab:further}
\begin{tabular}{l|cc|cc}
\hline
\multirow{2}{*}{Training}          & \multicolumn{2}{c|}{LIVEC}                               & \multicolumn{2}{c}{KonIQ-10k} \\
               & SRCC&PLCC                                                & SRCC&PLCC                       \\
\hline
DGDS           & 0.6958	&0.6902	 &0.6810	&0.6866 \\
GDS             & 0.7134	&0.7254  &\textbf{0.7255}	&0.7275 \\
\hline
LIVEC          &-       &-       &0.7074    &\textbf{0.7539} \\
KonIQ-10k      & \textbf{0.8001}  &\textbf{0.8096}  &-         &-      \\
\hline
\end{tabular}
\end{table}

\section{Conclusion}
The annotation of BIQA is labor-intensive and time-consuming, especially for authentic images. Training on synthetic data is expected to be beneficial, but synthetically trained models often suffer from poor generalization in authentic distortion datasets due to domain gaps. We make a key observation that the use of a larger synthetic distortion dataset may not improve or even be harmful to the performance of model generalization to authentic distortions, and analyze the reasons for this in the context of some research about domain adaptation.
To solve this challenge, we analyze how to measure the distance of source and target domains for BIQA under unsupervised scenarios and accordingly propose a distortion-guided unsupervised domain adaptation framework that reduces the negative transfer of the outlier source domains to the target domain and improves the generalization ability of the model.
Our approach significantly improves the model performance with a significantly smaller amount of training data in the settings of synthetic distortion to authentic distortion and synthetic distortion to algorithmic distortion. Besides, since our method is a sample-based approach, it is fully compatible with existing model-based approaches. Finally, we explore the upper bound of the performance of the method based on source domain selection on KADID-10k for generalizing to authentic distortion datasets and experimentally demonstrate the feasibility of generalizing to authentic distortions by generalizing over synthetic distortions. 

\section{Acknowledgment}
This work was partially supported by the Fundamental Research Funds for the Central Universities (QTZX23038), Postdoctoral Fellowship Program of China Postdoctoral Science Foundation (GZC20232034) and Postdoctoral Research Grant of Shaanxi Province (2023BSHEDZZ165).
{
    \small
    % \bibliographystyle{ieeenat_fullname.bst}
    % \bibliography{main}

}

% WARNING: do not forget to delete the supplementary pages from your submission 
% \input{sec/X_suppl}

\end{document}